\documentclass[publish,JRM,paper]{jaciiiarticle}
\usepackage[dvips]{graphicx} 
\usepackage{epsfig}  
\usepackage{multirow}

\begin{document}

\pagestyle{jaciiistyle}

\title{Tracking and Visualizing Signs of Degradation for an Early Failure Prediction of a Rolling Bearing}
\author{Sana Talmoudi*, Tetsuya Kanada**, Yasuhisa Hirata*}
\address{*Department of Robotics, Graduate Faculty of Engineering, Tohoku University,\\ 6-6-1 Aramaki Aza Aoba, Aoba-ku, Sendai, Miyagi, Japan. 980-8579\\
         E-mail: \{s.talmoudi,hirata\}@srd.mech.tohoku.ac.jp\\
         **D'isum Inc.,\\
         3-10-18 Takanawa, Minato-ku, Tokyo, Japan, 108-0074\\
         E-mail: kanada@d-isum.com}
\markboth{Talmoudi S., Kanada T. and Hirata Y.}{Tracking and Visualizing Signs of Degradation for an Early Failure Prediction of a Rolling Bearing}
\maketitle

\begin{abstract}
\noindent Predictive maintenance, i.e. predicting failure to be few steps ahead of the fault, is one of the pillars of Industry 4.0. An effective method for that is to track early signs of degradation before a failure happens. This paper presents an innovative failure predictive scheme for machines. The proposed scheme combines the use of full spectrum of the vibration data caused by the machines and data visualization technologies. This scheme is featured by no training data required and by quick start after installation. First, we propose to use full spectrum (as high-dimensional data vector) with no cropping and no complex feature extraction and to visualize data behavior by mapping the high dimensional vectors into a 2D map. We then can ensure the simplicity of process and less possibility of overlooking of important information as well as providing a human-friendly and human-understandable output. Second, we propose Real-Time Data Tracker (RTDT) which predicts the failure at an appropriate time with sufficient time for maintenance by plotting real-time frequency spectrum data of the target machine on the 2D map composed from normal data. Third, we show the test results of our proposal using vibration data of bearings from real-world test-to-failure measurements provided by the public dataset, the IMS dataset.
\end{abstract}


\section{Introduction}
Ball or rolling element bearing are used in the majority of the electrical machines \cite{ref1} as well as in electromechanical and mechanical systems. According to Bonnett et al. in \cite{ref2}, bearing failures is almost two-thirds of all motor failures. Being a key element in rotary machines installed in plants and factory, a bearing fault can be of a catastrophic effect in terms of safety and production cost. Therefore, an effective method for tracking the early signs of degradation and predicting the failure before it happens is a mandatory task.\par
A bearing, as any other mechanical element, while being subject to a deterioration, would exhibit a change in the vibration, in both amplitude levels as well as the spectral distribution. Therefore, vibration sensors have been widely used to obtain bearing monitoring data.\par
Most of the existing works on bearing condition monitoring are usually applied and validated on fault-seeded datasets. Classification and detection ability are validated. However, it remains challenging to apply those methods on a real-world-like situation for tracking the signs of degradation where the degradation is highly non-linear. Other works have been tested on real-world like situations using run-to-failure experiments. Two main dataset provided measurements from run-to-failure experiments and focused on the bearing faults: the dataset from the Intelligent Maintenance Systems (IMS), University of Cincinnati \cite{ref5} that has been published on the website of the Prognostic Data Repository of NASA \cite{ref6} and the bearing dataset by the Franche-Comt\'e Electronics Mechanics Thermal Science and Optics–Sciences and Technologies (FEMTO) institute \cite{ref7} that has been shared in the IEEE international conference of PHM 2012 for prognostic challenge. \par
On these run-to failure datasets, for diagnosis purpose, good results have been achieved on classifying the data into several classes for the different bearing conditions \cite{ref5,ref13,ref23,ref20,ref30,ref33,ref17,ref18,ref34,ref16,ref35}. Diagnosis only wouldn’t be of a use in a real-life like situation for failure prediction. A tracking method and an online prognosis solution is required. Therefore, another direction of research existed where several works \cite{ref21,ref12,ref27,ref19,ref26,ref14,ref29,ref38,ref36,ref37,ref22,ref15,ref24,ref25,ref28,ref31,ref32} have worked on tracking bearing faults, estimating certain indexes to predict the condition of the bearing. However, detection of early stage degradation and developing an effective real-time unsupervised tracker for bearing conditions remain challenging. The sensor signal exhibits a change in the amplitudes and\slash or change in the frequency response when the bearing is facing a deterioration, although, early signs of degradation can be overlooked. Additionally, initial mounting conditions would affect the lifetime and the degradation timeline of the bearing and would result in a new degradation pattern. The bearing might as well exhibit a self-healing phenomenon \cite{ref5} which would make the sensor data confusing for certain methods. Due to the these challenges, both data-driven and model-based conventional methods can be limited to cover effectively early signs of degradation and track them. Choosing a specific set of features will always limit the number and the types of faults to be detected and would not provide a generic method that can always cover new faults patterns.\par
In this paper we are proposing the use of the full range of the frequency components of the vibration data as dimensions without extracting any specific features because it is unknown how and where the first sign of degradation would appear in the frequency spectrum. We then define the similarity between the high-dimensional data instants and use a visualization method to plot the multi-dimensional space into a two-dimensional plane. Therefore, despite using all the original number of dimensions to analyze the data to avoid overlooking any possible meaningful information, we still can provide an human-friendly output that is easy to interpret. In this work, we are focusing on data visualization methods that are quite sensitive to detect very early degradation appearing as a faint change in frequency characteristics as well as vibration amplitude. Therefore, we are testing our analysis scheme with both visualization methods that proved to fulfill such characteristic: toorPIA \cite{ref10} and t-SNE \cite{ref11}. We are comparing the obtained results to determine which of the two visualization techniques is capable to effectively express the degradation pattern.\par
Additionally, we are proposing a Real-Time Data Tracker (RTDT) that can be used for tracking the bearing condition and therefore, predict the failure before it happens. Previous works tended to the estimation and prediction of Remaining Useful Life (RUL) based on learnt and trained models and classifiers. In this work, the RTDT is not estimating the RUL nor a health index based on learnt data or predefined models but instead tracks and visualizes the data from the machine in real-time issuing an alert at an appropriate time.\par
We are applying our proposal on the IMS bearing dataset to verify and validate our hypothesis. The IMS bearing dataset has been widely used for bearing diagnosis and prognosis and it presents a high challenge since it covers run-to-failure tests. During the tests, even though similar bearings have been installed, the bearing behaviors were different in type and timeline of the exhibited faults. Therefore, the use of this specific dataset will be the ultimate test for our method to prove the ability of classifying, detecting and tracking the signs of degradation without prior training or complex modeling. Moreover, by validating our method on the IMS dataset and covering the multiple reported faults, we can prove that our method is generic. The same process can be used on any type of facility\slash component for the detection and prediction of unknown and unexperienced events, which is the case of a real-world like situation where events are highly non-linear and sensor signals are non-stationary.\par
The novelty of the proposed solution is the combination of the use of the full frequency spectrum with a data visualization method for a real-time failure prediction. By the work presented in this paper, our contributions are as follow: (1) effectively classify and identify the different bearing conditions, (2) detect early signs of degradation to achieve failure prediction, (3) differentiate between types of failure (4) propose a real-time machine monitoring solution and (5) do not require any training data sets and can be used within a week after installation. Additionally, we are providing an output that is easy to show the tracking of signs of failure so that an operator even without any complex data science background can use our solution to take the appropriate action in the appropriate time. scheme does not require \par
The rest of this paper will start with a digest of the IMS dataset in section 2. In section 3, we are detailing our analysis method. Section 4 is focusing on the application of our analysis scheme on the IMS dataset to classify and identify the bearing conditions, to track the early signs of degradation and to visualize multiple faults that have been exhibited by one of the bearing during one of the tests. In section 5, we are proposing the real time data tracker, RTDT, and applying our proposal for a real-time tracking of the bearing conditions from IMS dataset. Section 6 is dedicated to further discussion on the obtained results and positioning our work among related works. Finally, we are concluding our work in section 7.

\section{Digest of the IMS Bearing Dataset}

The dataset was provided by IMS, University of Cincinnati \cite{ref5} and shared on the the website of the Prognostic Data Repository of NASA \cite{ref6}. The dataset consists of several bearing run-to-failure tests that were conducted on a specially designed test rig.\par
The test rig consisted of four test bearings mounted on one shaft. The shaft was driven by an AC motor and coupled by rub belts \cite{ref5}. A radial load of 6000 lbs was applied to the shaft and bearing via a spring mechanism and the bearings were then under normal load conditions. The rotation speed was kept constant and equal to 2000 rpm. Lee et al. mentioned in \cite{ref6} that the designed lifetime of the bearings is more than 100 million revolutions. Additionally, a magnetic plug was installed in the oil feedback pipe collecting the metallic debris and to serve as an evidence of bearing degradation and to cause an electrical switch to close when the debris exceed certain threshold \cite{ref5}.\par 
The bearings used in this test rig are Rexnord ZA-2115 double row bearings.To measure the vibration signals from the bearings, PCB 353B33 High Sensitivity Quartz ICP\textregistered accelerometers were attached to the bearing housing. During test 1, two accelerometers were attached for each bearing to cover x-axis vibration as well as y-axis vibration. In the rest of the tests, only one accelerometer was attached per bearing. Data acquisition was performed by a National Instruments DAQCard-6062E data acquisition card with a sampling rate of 20 kHz. Each recorded measurement covered 1 second of the bearing operation, at 10-minutes intervals (except the first 43 measurements of test1) and contained 20480 data samples. The data collection was conducted by a LabVIEW based-program.\par
In their paper \cite{ref5}, Qiu et al. stated that the experiments of test 1 led to: an inner race defect in bearing 3 and a mixture of roller defect and outer race defect in bearing 4. In \cite{ref6} Lee et al. stated that by the end of test 2, bearing 1 exhibited an outer race failure. Further details on the bearings characteristics can be found in \cite{ref5,ref6}.

\section{Methodology}

\begin{figure}[t]
\centering
\includegraphics[width=7cm]{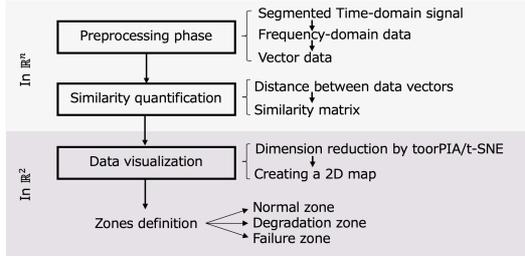}
\caption{Process flow of the proposed analysis method }
\label{fig4}
\end{figure}

In our previous work \cite{ref3}, we proposed the backbone of our analysis method for machinery failure prediction. Our proposal consists of 3 main phases: (i) a preprocessing phase to generate the high-dimensional vector data from the raw data, (ii) similarity quantification on the high-dimensional data vectors then (iii) data visualization of the high-dimensional vector data in a two-dimensional plane. On the two-dimensional plane, and based on the distribution of the data points, we can define the multiple machine condition zones: the safe zone, the degradation zones and the failure zone. The process flow of the analysis method is detailed in \textbf{Fig.\ref{fig4}}.
In the next sections we will use the following notations:
\begin{itemize}
\item We refer to a vector \textit{U} from the origin in the multi-dimensional space by \textit{\textbf{U}}, where  $\bm{U}\in\Re^n$
\item We refer to a vector \textit{U} from the origin in the two-dimensional plane by \textit{\textbf{u}}, where $\bm{u}\in\Re^2$
\item We refer to the vector formed by a point A and a point B in the multi-dimensional space by $\overrightarrow{AB}$
\item We refer to the vector formed by a point A and a point B in the two-dimensional plane by $\overrightarrow{ab}$
\item We refer to the inner product between a vector U and a vector V by $\langle \bm{U},\bm{V} \rangle$
\end{itemize}

\subsection{The Preprocessing Phase}

Sensor data is most often collected as time-domain data.
We start by dividing the time-domain data into segments. Next, on each segment, we apply a windowing function, the Hanning window, and convert the time-domain data into the frequency-domain using the Fast Fourier Transform (FFT), more specifically, the Discrete Fourier Transform (DFT). The FFT result being symmetric, we only consider the positive frequencies. We then use the obtained spectrum to generate a multidimensional vector data \textit{\textbf{V}}. In certain cases where the sampling rate is high, we smoothen the spectrum to reduce the white noise.\par
In our method we do not extract any specific feature, but instead we consider the whole spectrum as a high-dimensional data because it is unknown how and where in the frequency spectrum signs of failures and early degradation might be hidden. Extracted features are generated by equations relating the original time-domain data and\slash or frequency-domain data. Despite that effective features have been extracted in previous works and used to describe the machine condition, the same set of extracted features that was successful in a fault scenario A couldn’t illustrate effectively a fault scenario B. Therefore, by considering the whole spectrum, we are not limited to a specific optimized set of extracted features and we can avoid overlooking any important information that the sensor data signal holds. \par
The preprocessing phase then results in transforming the raw data to a set of high dimensional data vectors $\{\bm{V_i}\}_{1\leq i \leq N}$ where $\bm{V_i}\in\Re^n$ and $\bm{V_i}=(v_{i1},v_{i2},\cdots,v_{in})$, N is the number of data vectors and n being the number of dimensions, i.e. frequency components of the spectrum.

We use the distance in the multi-dimensional space given by  \textbf{(\ref{eq2})}, to quantify the similarity between a data vector $\bm{V_i}$ and a data vector $\bm{V_j}$.

\begin{equation}
d_{ij} = \sqrt{\langle \bm{V_i},\bm{V_i} \rangle + \langle \bm{V_j},\bm{V_j} \rangle - 2 \langle \bm{V_i},\bm{V_j} \rangle}
\label{eq2}
\end{equation}

Where:
\begin{itemize}
\item $\bm{V_i},\bm{V_j}\in\Re^n$ $\forall i,j\in[1,N]$ 
\item $\langle \bm{V_i},\bm{V_j} \rangle$ is the inner product of $\bm{V_i}$ and $\bm{V_j}$
\end{itemize}
From the obtained distance values between all data vectors, we define then a similarity matrix, \textit{S}. \textit{S} is an $n \times N$ matrix expressed by \textbf{(\ref{eq3})}, where, n is the number of dimensions, i.e. frequency components of the spectrum, N is the number of data vectors and $d_{ij}$ is the distance in multidimensional space between a data vector $\bm{V_i}$  and a data vector $\bm{V_j}$

\begin{equation}
S = \begin{pmatrix}
d_{11} & d_{12} & \cdots & d_{1n}\\
d_{21} & d_{22} & \cdots & d_{2n}\\
\vdots & \vdots & \ddots & \vdots\\
d_{N1} & d_{N2} & \cdots & d_{Nn}
\end{pmatrix}
\label{eq3}
\end{equation}

\subsection{Data Visualization}

According to the distance between data, the high-dimensional data vectors are mapped in a two-dimensional plane using a dimension reduction technique such as: toorPIA \cite{ref10} or t-SNE \cite{ref11}. For each $\bm{V_i}\in\Re^n$, we assign then $\bm{v_i}\in\Re^2$ where $\bm{v_i}=(v_{ix},v_{iy})$. \par
Dimensional reduction schemes tend to translate the similarity seen in the multidimensional space to clusters in the two-dimensional plane. The aim of dimensionality reduction is to preserve as much of the significant structure of the high-dimensional data as possible in the low-dimensional map \cite{ref11}. The differences among the existing visualization methods relies in the definition of the significant structure to preserve. Non-linear techniques such as t-SNE and toorPIA tend to preserve the structure of the similar data, so known as the local structure. However, in certain applications, it is desirable to preserve the far structure to some degree. Then the choice of dimensional reduction schemes depends on the case according to the appropriate balance between preserving the local and the far structure.\par
In this work, we are testing both visualization tools, t-SNE and toorPIA, to determine which method is more appropriate to be used for tracking the signs of degradation to achieve an effective prediction.\par
From the distribution of the data points on the two-dimensional plane, we can define the different states the machine\slash component has been through by defining the different zones: (1) the safe zone corresponding to the early data covering usually the normal and possible very light degradation, (2) the degradation zones assuming that there might be several levels and types of degradation and (3)the failure zone. Moreover, the trajectory formed by the movement of the datapoints on the two-dimenional plane is the key to track the evolution in the machine\slash component condition.

\section{Classification of Bearing Condition and Visualization of Failure Signs in the IMS Bearing Dataset}
In this section, we are applying our proposed analysis method on the IMS dataset to verify that our analysis scheme permits (a) the classification the bearings conditions, (b) tracking of the change in the bearing condition, (c) that very early stage of degrading data to be distinguished from normal data and (d) that multiple failure case can be visualized differently from single failure case. \par
To verify these features and to cover all the reported types of faults in the dataset, we are using all the measurement files data from test 1 and from test 2 data as follow: we refer by B1T2 to the case where we consider the data recorded on channel 1 \cite{ref6} during \textbf{T}est \textbf{2}. This data covers the vibration of \textbf{B}earing \textbf{1} condition evolution from healthy to Outer Race Failure (ORF). B3T1x refers to the case where we consider the data recorded on channel 5 \cite{ref6} during \textbf{T}est \textbf{1}. This data covers the vibration on the \textbf{x}-axis of \textbf{B}earing \textbf{3} condition evolution from healthy to Inner Race Failure (IRF). We refer by B4T1x to the case where we consider the data recorded on channel 7 \cite{ref6} during \textbf{T}est \textbf{1}. This data covers the vibration on the \textbf{x}-axis of \textbf{B}earing \textbf{4} condition evolution from healthy to Roller Defect (RD) mixed with Outer Race Failure (ORF).

\subsection{The Preprocessing Phase}

\begin{figure}[t]
\centering
\includegraphics[width=7cm]{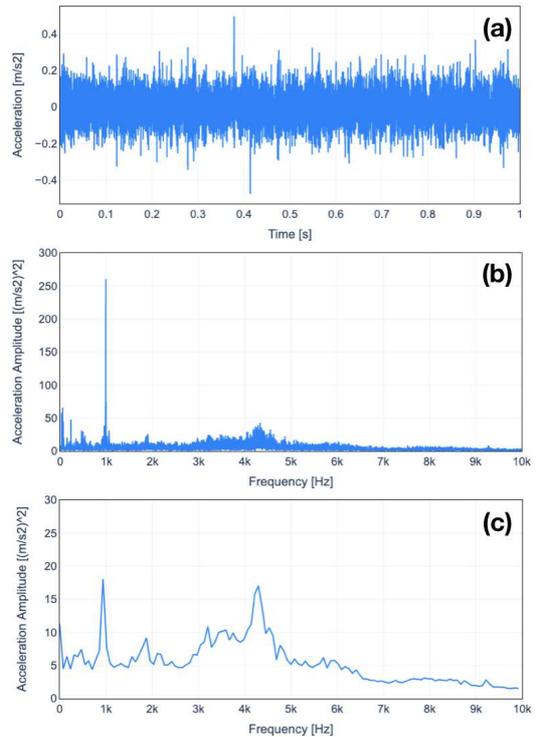}
\caption{Data signal of measurement 2004.02.13.15.52.39 through the preprocessing phase (a) time-domain of the data segment (b) original spectrum (c) smoothed spectrum}
\label{fig5}
\end{figure}

\begin{table}[t]
 \begin{center}
   \caption{Summary of the vector sets at the end of the preprocessing phase}
	   \begin{tabular}[t]{c|c c} 
	   \hline
	  & No. vectors & No. dimensions\\
	   \hline
	   B1T2 & 984 & 128\\
	   B3T1x & 2156 & 128\\
	   B4T1x & 2156 & 128\\
	   \hline
	   \end{tabular}
   \label{table2}
 \end{center}
\end{table}

In the current work, we are considering each of the 1-second-long measurement as a data segment. A data segment then consists of 20480 vibration data (\textbf{Fig.\ref{fig5}.a}). We generate the spectrum of each data segment using FFT, more specifically DFT. We consider the positive frequency components and we obtain a spectrum of size $2^{14}$ (\textbf{Fig.\ref{fig5}.b}). We then smoothen the obtained spectrum to reduce the white noise effect. For this purpose, we compute the average of every N consecutive frequency components and obtain a smoother spectrum. The frequency resolution is reduced from 10kHz⁄16384 Hz to 10kHz⁄16384/N Hz. N should be chosen to be an appropriate number considering the frequency characteristics of the vibration. In the following data processing, the number of 128 was chosen as N (\textbf{Fig.\ref{fig5}.c}). The $2^7$ frequency components form the dimensions in our multidimensional space. Therefore, by the end of the preprocessing phase, every 1s-data segment is converted to a data vector of $2^7$ dimensions. \textbf{Table \ref{table2}} summarizes the output of the preprocessing phase for each of the cases, B1T2, B3T1x and B4T1x.\par
The parameters for the data segment length and the frequency resolution are directly related to the nature of the original vibration signal. Such parameters can be different for other application on other monitored systems, although, the core preprocessing remains the same.

\subsection{Similarity Quantification and Data Visualization}
In this stage we are performing the dimension reduction from $\Re^n$ to $\Re^2$. We are mapping the high-dimensional data vectors in a two-dimensional plane using a dimensional reduction technique.\par
In this paper, we are testing our analysis scheme with both reduction techniques: toorPIA and t-SNE. Both methods have been proved to be sensitive to small changes in the data which is a key characteristic to detect early sign of degradation. We were provided with an implementation of toorPIA from Toor Inc. \cite{ref10} and we used the t-SNE implementation from Scikit \cite{ref9}. \par
The position of the high-dimensional data vectors in the two-dimensional plane is defined based on multidimensional distance between them. Therefore, to use toorPIA, we first quantified the similarity between the high-dimensional data vectors and generated the \textit{S} matrix based on \textbf{(\ref{eq2})}. \textit{S} matrix is our input to toorPIA mapping function. However, from \textbf{(\ref{eq2})}, \textit{S} matrix can be simplified to an upper triangular matrix with a null diagonal, given the following characteristics: (a) $d_{ij} = d_{ji}$ and (b) $d_{ii} = 0$\par
On the other hand, to use the Scikit implementation of t-SNE, we use the set of the generated data vectors as input to the tsne function. In the Scikit implementation, the similarity between the vectors is performed within the tsne function. However, we can specify the metric to be used to define the similarity between the data vectors. In order to effectively compare the output of toorPIA and t-SNE, we set the metric to be the function given by \textbf{(\ref{eq2})}. Moreover, when using t-SNE, we tried different perplexity values \cite{ref11}: 10, 30 and 50 \cite{ref11}.

\subsection{Experimental Results and Discussion}

\begin{figure}[t]
\centering
\includegraphics[width=7cm]{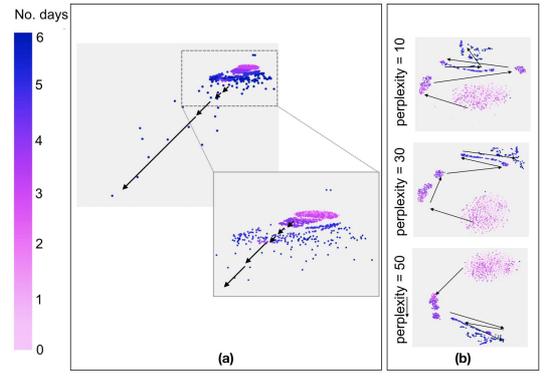}
\caption{Movement of the data points in the two-dimensional plane from the data of B1T2 using for visualization (a) toorPIA and (b) t-SNE}
\label{fig6}
\end{figure}

\begin{figure}[t]
\centering
\includegraphics[width=7cm]{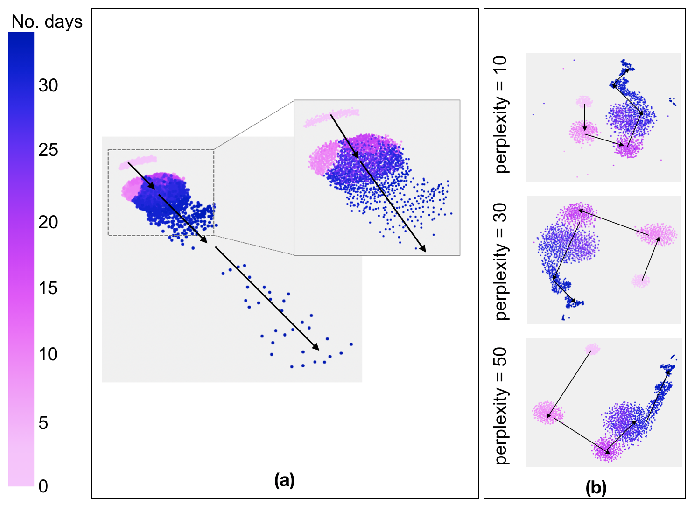}
\caption{Movement of the data points in the two-dimensional plane from the data of B3T1x using for visualization (a) toorPIA and (b) t-SNE}
\label{fig7}
\end{figure}

\begin{figure}[t]
\centering
\includegraphics[width=7cm]{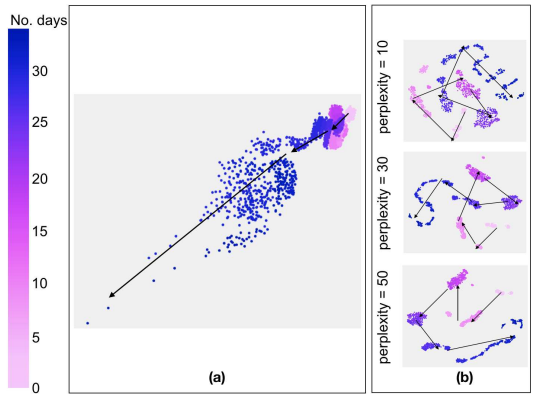}
\caption{Movement of the data points in the two-dimensional plane from the data of B4T1x using for visualization (a) toorPIA and (b) t-SNE}
\label{fig8}
\end{figure}

The obtained two-dimensional maps for B1T2, B3T1x and B4T1x are shown in \textbf{Fig.\ref{fig6}}, \textbf{Fig.\ref{fig7}} and \textbf{Fig.\ref{fig8}}, respectively with both, toorPIA and t-SNE.

\textit{\textbf{Classification and identification:}}\par
In both, toorPIA and t-SNE results, and for all the 3 fault cases, the degradation of the bearing is clearly visualized by using our analysis scheme. Several condition states/classes were presented by different clusters on the 2D map. The early stage of degradation can be detected from the data visualization. As we advance in time, new clusters for the bearing’s conditions appear. This is explained by the fact that degradation increases over time until reaching the failure instant.\par
t-SNE, same as toorPIA, identified and classified the data into 4 main condition classes: healthy, light degradation, severe degradation and failure for both B3T1x and B1T2 cases. As for B4T1x case, Qiu et al. \cite{ref5} mentioned that bearing 4 exhibited a mixture of outer race defect and roller defect. On the toorPIA map, we can see that the data clearly distributed on 4 main areas corresponding to the 4 main condition classes. However, from the t-SNE maps with low perplexity value, it is difficult to tell between degradation class, failure 1 class and failure 2 class. By increasing the perplexity, we can see that the data show distribution in 4 main states of the bearing. t-SNE is sensitive to the close distances (to preserve the local structure) while not much to far distances although the sensitivity depends on the perplexity (this can be seen by the comparison of the results with the different perplexity values, 10, 30 and 50).\par
We then can conclude, that for all the faults scenarios, B1T2, B3T1x and B4T1x, with both toorPIA and t-SNE, we were able to differentiate between the main condition states.

\textit{\textbf{Tracking:}}\par
To verify the ability of our proposed method to track signs of degradation, we plotted the trajectory of how the data points moved along the two-dimensional maps as shown in \textbf{Fig.\ref{fig6}}, \textbf{Fig.\ref{fig7}} and \textbf{Fig.\ref{fig8}}, for B1T2, B3T1x and B4T1x respectively. Compared to t-SNE, toorPIA mapping showed an ability of tracking the failure, especially in the case B4T1x, where the bearing exhibited a mixture of two faults, RD and ORF. For t-SNE maps, the inter-clusters distances in the low-dimensional space are not reflecting the distances between those clusters in high-dimensional space and therefore not tracking the evolution in the bearing condition. In other words, from t-SNE maps, we cannot have an appropriate feedback on the condition of the bearing or on how severe the situation is. toorPIA map had linear structure with the change in the bearing condition. On the maps generated by toorPIA, the data distributed according to the degradation level, in a specific direction. Due to the difference in their design philosophy, t-SNE is more appropriate for data classification while toorPIA is more appropriate for tracking the degradation degree and provide a more natural output for human understanding. It is important to mention that the first application of t-SNE was to visualize and classify data \cite{ref11}. Based on the findings on the tracking ability of toorPIA and t-SNE, in this section, we are moving forward with the maps obtained by toorPIA only.

\begin{figure}[t]
\centering
\includegraphics[width=7cm]{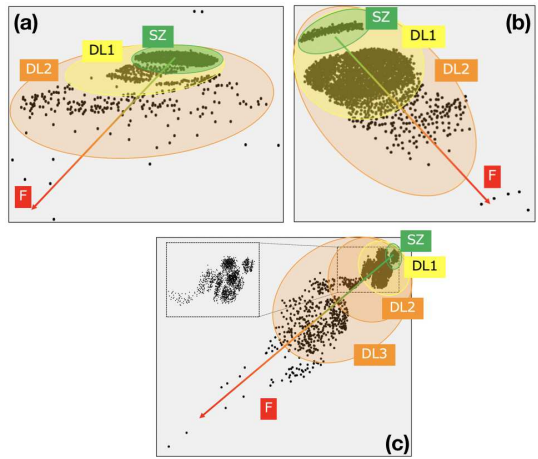}
\caption{Defining the major condition zones for (a) B1T2, (b) B3T1x and (c) B4T1x}
\label{fig9}
\end{figure}

\begin{figure}[t]
\centering
\includegraphics[width=7cm]{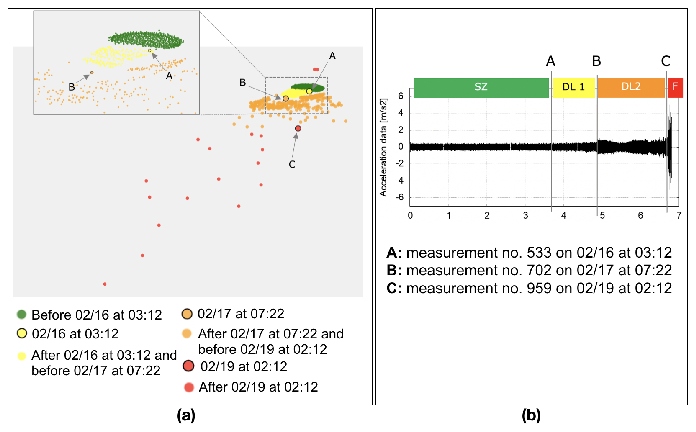}
\caption{Tracking of the change in the bearing condition in the case of B1T2 (a) the start of each condition fby toorPIA map, (b) the corresponding time stamps}
\label{fig10}
\end{figure}

\begin{figure}[t]
\centering
\includegraphics[width=7cm]{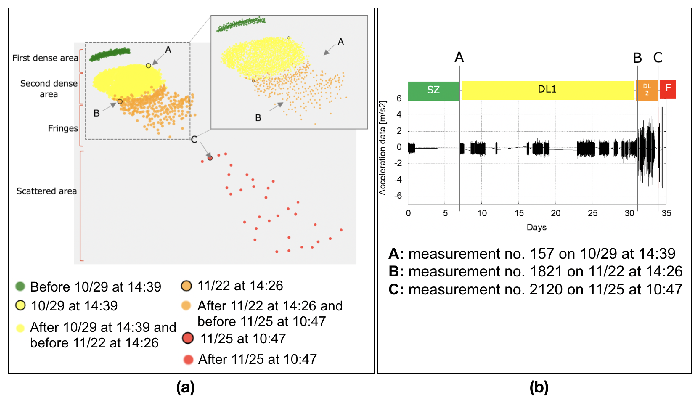}
\caption{Tracking of the change in the bearing condition in the case of B3T1x (a) the start of each condition by toorPIA map, (b) the corresponding time stamps}
\label{fig11}
\end{figure}

\begin{figure}[t]
\centering
\includegraphics[width=7cm]{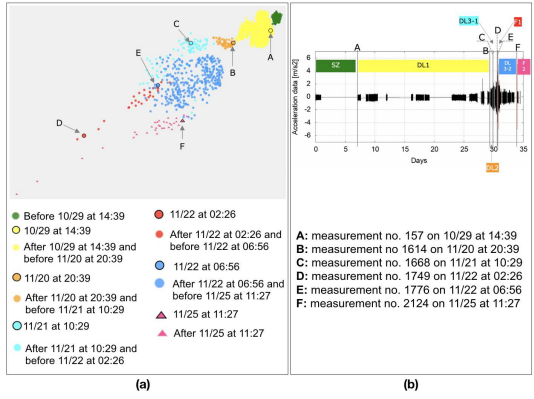}
\caption{Tracking of the change in the bearing condition in the case of B4T1x (a) the start of each condition by toorPIA map, (b) the corresponding time stamps}
\label{fig12}
\end{figure}

\textit{\textbf{Zones definition:}}\par
For each of the three faults scenarios, we are then defining the zones on the two-dimensional map, corresponding to the identified main condition levels.\par
\textit{B1T2} (\textbf{Fig.\ref{fig9}.a}): the dense cluster corresponds to the healthy data. As the degradation starts, the data drift out of the dense cluster in fringes. As the degradation worsen, the data separate from the early degradation area forming another degradation level area. When the failure happens, the data move further from the healthy and degradation areas. Moreover, the fringe-way of the distribution of the data suggests that there have been sub-levels of each degradation level. We then define 4 zones assuming that, in test 2, bearing 1 went through 4 main stages: (1) Safe zone (SZ) = healthy stage, (2) Degradation level 1 (DL1) = early degradation, (3) Degradation level 2 (DL2) = severe degradation stage and (4) Failure (F) = failure stage\par
\textit{B3T1x} (\textbf{Fig.\ref{fig9}.b}): the small dense cluster on the top of the map corresponds to the early data and therefore to the healthy data. While the next dense cluster is early light degradation. The gap between the two dense clusters is actually due to a 6-days gap in measurements. 6 days are 18 million revolutions which is 18\% of the bearing expected lifetime given by the authors in \cite{ref6}. Therefore, we assume that some change in the bearing condition has happened and resulted in the data points not overlapping with the first hours data that formed the top dense cluster. If the missing 6-days data would have been available, the data would continuously distribute as in B1T2 results. The sparser points that followed the dense area correspond to the severe degradation. These points distribute in fringes. We assume that there are sublevels of the serious degradation conditions. As the degradation becomes more serious, the data keep drifting away from the dense areas and become sparser. When failure started to happen, the data was extremely far from the healthy and degradation data. We then define 4 zones assuming that, in test 1, bearing 3 went through 4 main stages: (1) Safe zone (SZ) = healthy stage,  (2) Degradation level 1 (DL1) = early degradation, (3) Degradation level 2 (DL2) = severe degradation stage and (4) Failure (F) = failure stage.\par
\textit{B4T1x} (\textbf{Fig.\ref{fig9}.c}): the small dense cluster on the top right of the map correspond to the early data and therefore to the healthy data. The next dense clusters on the left side of the small dense cluster correspond to the first stage of degradation: the light degradation. The data of this area is divided into 3 sub-clusters. This may be explained by the fact that according to Qiu et al. \cite{ref5}, in Test 1, Bearing 4 exhibited two types of faults: RD and ORF. There was no information about the way the mixture of the two faults happened. At early stage of degradation, the state of the bearing was sometime dominated by degradation of type 1 and sometime dominated by the degradation type 2 and sometime witnessed both degradation types at equal intensity at the same time. However, the level of the degradation was still light. As the degradation became more serious, the data kept drifting away from the dense areas while getting sparser. When the failure started to happen, the data became completely far from the healthy and degradation data. However, the data returned to the degradation zone while occupying a different side of the degradation area. This might be explained by the following: the degradation level 3 (DL3) as shown in \textbf{Fig.\ref{fig12}} , led to a failure F1 (it can be the Roller Defect or Outer Race Defect). Then the bearing had a self-healing phase as mentioned by Qiu et al. \cite{ref5}. However, the second type of degradation worsened and dominated the degradation condition of bearing 4. Consecutively, the bearing witnessed the second type of failure, F2 and the data drifted again from the degradation area, in a different direction from F1. Therefore, we assume that in test 1, bearing 4 has been through 7 stages: : (1) Safe zone (SZ)= healthy stage, (2) Degradation level 1 (DL1) = early degradation stage, (3) Degradation level 2 (DL2) = medium degradation stage, (4) Degradation level 3, phase 1 (DL3-1) = the stage of severe degradation, phase 1, (5) Failure type 1 (F1) = the stage of the 1st type of failure, (6) Degradation level 3, phase 2 (DL3-2) = the stage of severe degradation, phase 2 and (7) Failure type 2 (F2) = the stage of the 2nd type of failure.\par
The above definition of the zones corresponding to the bearing condition in B1T2, B3T1x and B4T1x, can be verified by the temporal distribution of the data shown in \textbf{Fig.\ref{fig6}}, \textbf{Fig.\ref{fig7}} and \textbf{Fig.\ref{fig8}}, respectively. \par
Within each degradation level, the data distributed into fringes. Every fringe corresponds to a sub-level of the degradation level. Each failure\slash defect scenario has different number of fringes. As mentioned in section 1, this can be explained by the fact that the evolution of each failure\slash defect scenario has different timeline and pattern. Therefore, our method was capable of visualizing these differences and adapt to how discrete the degradation sub-levels can be. Model-based method or data-driven method using specific extracted features might overlook some intermediate sub-levels of degradation, that sometimes can be of a high importance to predict a fatal failure before it happens.\par
Additionally, from the results of B4T1x, we were able differentiate the two types of failures that Bearing 4 exhibited in the same test, as mentioned by Qiu et al. \cite{ref5}. In B1T2 and B3T1x, where the bearing, according to \cite{ref5,6}, exhibited only one type of failure, the result maps were mono-direction. On the other hand, in B4T1x, the result map was bi-directional. However, the validation of which of the different faults is F1 or F2 can be only possible if any further information about the test timeline is provided.\par
Consequently, we can confirm that our method can effectively track the levels and types of degradation.

\begin{table}[t]
 \begin{center}

   \caption{Summary of the faults timeline}
	   \begin{tabular}[t]{c|c c c} 
	   \hline
	    & B1T2 & B3T1x & B4T1x\\
	   \hline
	   A & 533 & 157 & 157\\
	   B & 702 & 1821 & 1614\\
	   C & 959 & 2120 & 1668\\
	   D & - & - & 1749\\
	   E & - & - & 1776\\
	   F & - & - & 2124\\
	   \hline
	   \end{tabular}
   \label{table3}
 \end{center}
\end{table}

\begin{table}[t]
 \begin{center}
   \caption{Comparison of how early degradation can be identified}
	   \begin{tabular}[t]{c|c c c} 
	   \hline
	   Work & B1T2 & B3T1x & B4T1x\\
	   \hline
	   Qian et al. \cite{ref28} & 583 & 1808 & -\\
	   Ben Ali et al. \cite{ref15} & 553 &1561 &1560\\
	   Hasani et al. \cite{ref31} &547 &2027 &1641\\
	   Our method &533 &1821\footnotemark &1614\footnotemark[\value{footnote}]\\
	   \hline
	   \end{tabular}
   \label{table4}
 \end{center}
\end{table}
\footnotetext{We are considering point “B” for comparison with other works, since we cannot rely on point “A” due to the gap in the measurements in test 1 }

\textit{\textbf{First signs:}}\par
 On the obtained 2D maps, we identified the start of every condition stage each bearing has been through by tracking the data movements. \textbf{Table \ref{table3}} summarizes the start of every identified condition for each bearing.\par
\textit{B1T2:} in \textbf{Fig.\ref{fig10}.a}, we were able to identify: the start of the early degradation by the point \textbf{A}, the start of the severe degradation by the point \textbf{B} and the start of the failure by point \textbf{C}. During the severe degradation state, the vibration amplitudes in the time-domain data increased and decreased. This is due to the pattern of fault as well as the `self-healing' phenomenon mentioned in literature. This can be tricky for conventional method. On the other hand, in our map, the data were not plotted in the normal zone, but instead plotted in degradation area. This can be explained by the fact that despite the decrease in amplitude in time-domain data, the spectral distribution is different from healthy data.\par 
\textit{B3T1x:} in \textbf{Fig.\ref{fig11}.a}, we were able to identify: the start of the early degradation by the point \textbf{A}, the start of the severe degradation by the point \textbf{B} and the start of the failure by point \textbf{C}. When the scattered data drifted out of the dense area, to make some fringes, this marked the start of more serious levels of the degradation. Point \textbf{B} marked the start of the fringe area. Only 21 out 299 points of what we assigned as DL2, were plotted at the edge of DL1 area resulting in 93\% of the data of DL2 to be correctly placed.\par
\textit{B4T1x:} in \textbf{Fig.\ref{fig12}.a}, we were able to identify: point \textbf{A} as the start of the early degradation stage, point \textbf{B} started the medium degradation level, point \textbf{C} started the stage of severe degradation level, phase 1, point \textbf{D} marked the start of the the failure type 1, while point \textbf{E} marked of the stage of severe degradation level, phase 2 and finally point \textbf{F} marked the start of the failure type 2.\par

\textbf{Table \ref{table4}} presents a comparison between our results and other works in terms of how early we identify degradation. In test 1, the recorded data had a 6-days gap in the measurements in the first week. This gap prevented the tracking of the actual start of the degradation, although, our method could place the data points at the right position in the two-dimensional space. \par	 
In this section, we confirmed that combining the use of the full spectrum and data mapping has potential to predict a failure since the vibration data at a very early stage of degradation could be distinguished from normal data on the map. The question is how to distinguish very early stage of degrading data from normal data given only normal data. In the real industrial situations, we need a real-time tracking of the data, being provided only with early normal data to monitor the bearing rather than an offline solution. Therefore, in the following section, we are proposing an online solution as a real time data tracking solution, the RTDT, for the monitoring of the bearing condition.

\section{Real-Time Data Tracker (RTDT) for the Prediction of Bearing Faults}

\subsection{Concept of RTDT}
\begin{figure}[t]
\centering
\includegraphics[width=7cm]{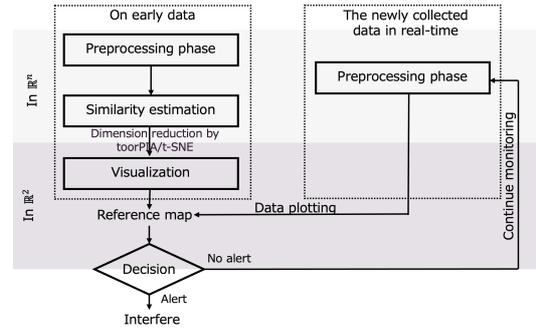}
\caption{Process flow of the RTDT }
\label{fig13}
\end{figure}
The RTDT relies on the same core technology of our proposed analysis scheme as detailed in section 3: the use of full spectrum data combined with data visualization. However, the process flow differs to provide a real-time solution. The process flow is given by \textbf{Fig.\ref{fig13}} and summarized by the following steps:\par
\textit{\textbf{Step 1:} Generating a Reference Map (RM)}\par 
We form a two-dimensional map using the early collected data and a \textit{zero-data} vector in multidimensional space. The early collected data represent the best condition for the machine\slash component. The choice of the length of the data used to generate the RM mainly depends on the machine\slash component to be monitored. We suggest the first 24h or 48h data to be considered for generating the RM. Most often, a machine\slash component should perform ideally during at least the first 48h. This duration can be reduced if the measurements were taken frequently in the first period where the machine was running. Additionally, we also define a \textit{zero-data} vector in the multidimensional space, defined by $\bm{O}=(0,0,\cdots,0)$, where $\bm{O}\in\Re^n$ where n is the number of dimensions. This \textit{zero-data} presents the origin of the coordinate system in the multi-dimensional space.\par
The RM is then formed, following the same steps described in section 3, by preprocessing the N data segments recorded in the early stage of the machine\slash component operation and quantifying the similarity and visualizing the N data vectors and the $\bm{O}$ vector.\par
\textit{\textbf{Step 2:} Plotting the newly collected data}\par
In a real-life like situation, the monitoring data will continue to be recorded, daily, every few minutes or every few hours. On the RM obtained after the first 48h, we add a newly acquired measurement. We start by running our preprocessing algorithm on the newly acquired measurement as a data segment to generate a test vector data $\bm{X}$. The test vector $\bm{X}$ will be then plotted on the RM. There are many approaches that can be implemented to plot the vector $\bm{X}$ into the RM. In the current work, we employed a geometry-based algorithm as described in Appendix A.\par
Therefore, for each newly recorded measurement, we define $\bm{x}\in\Re^2$ where $\bm{x}=(x_x,x_y)$. The data will first be plotted in the safe-zone (within the normal data area and its surroundings) on the reference map if the bearing condition is still normal. Then, as we progress in time, the degradation becomes more serious and the data will start drifting out of the safe-zone.\par
\textit{\textbf{Step 3:} Quantitative decision making}\par
For each test data vector $\bm{X}$, we define a waring factor $\rho$ as follow:\par
(i) From the reference map data in the two-dimensional space, we find $\bm{g}$, the center of gravity of the reference data in the two-dimensional plane\par
(ii) Next, from the reference map data in the two-dimensional space, we find $\bm{p}$ defined by $\langle \overrightarrow{gp} , \overrightarrow{gx} \rangle_{max}$ \par
(iii) Then we estimate $\rho$ where $\rho = \frac{\|\overrightarrow{gx}\|}{\|\overrightarrow{gp}\|}$\par
After collecting a number \textit{M} of test data vectors $\bm{X}$, we estimate the average value $\rho_{avg}$ from the \textit{M} calculated $\rho$ values. Using $\rho_{avg}$, we perform degradation assessment and decide whether or not to release a warning or alert. The number \textit{M} depends of how often we are recording data.\par
Based on its definition, $\rho$ is then the quantitative measure of how the newly acquired data moved compared to the distribution of the normal data rather than a model-based predicted value. Our method calculates $\rho$  using the actual position of the newly acquired data and based on the initial distribution of the normal data. Therefore, to set $\rho_{avg_{th}}$ as a threshold of $\rho_{avg}$, we do not need to train our method or have a prior knowledge of the whole distribution of $\rho$ for both normal and faulty states.\par
If the data is within the safe zone (early data area and its close surroundings), $\rho_{avg}\approx 1$. In this case, the bearing is still considered healthy. We are interested to be notified when the bearing already entered a light degradation level (to avoid unnecessary actions) and before the severe degradation comes to an end (to avoid failure). In other words, an alert should be issued within degradation level 1 and before degradation level 2 comes to an end. Based on this idea, we set the $\rho_{avg_{th}}$ to 2. It means the distance between the test data and $\bm{g}$, the gravity center of normal data, is about double of the distance between the normal data nearest from the test data and $\bm{g}$. If $\rho_{avg}$ value reaches $\rho_{avg_{th}}$, an alert is issued.\par
By combining the two-dimensional plane and the evolution graph of $\rho_{avg}$, we provide the operator with two possible visual outputs to help track the machine\slash component condition, minimizing false alerts as well as overlooking possible slight change of condition.\par
The core of the proposed RTDT scheme remains valid for the monitoring of different machines. However, a possible tuning and further optimization of the definition of the warning factor $\rho$.

\subsection{Application of RTDT for Bearing Faults Prediction in the IMS Dataset}

\begin{table}[t]
 \begin{center}
   \caption{No. vectors used  as reference\slash test to validate the RTDT on IMS dataset}
	   \begin{tabular}[t]{c|c c c} 
	   \hline
	   Case &Reference set &Test  set\\
	   \hline
	   B1T2 & 288 & 696 \\
	   B3T1x & 156  & 2000\\
	   B4T1x& 156  & 2000 \\
	   \hline
	   \end{tabular}
   \label{table5}
 \end{center}
\end{table}

The IMS dataset, consisting of run-to-failure tests, is a good dataset candidate to test the validity of the RTDT to predict bearing failures. To generate the RM, assuming that the bearings maintained a healthy condition during the first 48h, we use these early measurements, as reference set and the \textit{zero-data} vector as described in sub-section 5.1. In this study, we are presenting a primitive comparison between RM be generated by toorPIA and RM generated by t-SNE. The rest of the data in each test is considered as the test set. A summary of the use of the IMS dataset to validate the RTDT is given by \textbf{Table \ref{table5}}.\par
After RM generation, we iterate over the test set to plot each measurement on the RM. For each measurement from the test set, we estimate the waring factor $\rho$. In the current work, we are setting \textit{M} to 12 to estimate the average value of $\rho_{avg}$ which in the case of IMS dataset means that we are averaging $\rho$ every 3h. As mentioned in the sub-section 5.1, if $\rho_{avg}\approx 1$, the newly acquired data is being plotted within the safe zone (normal data area and its close surroundings). We set $\rho_{avg_{th}} = 2$. Once this threshold is crossed, an alert is issued. We are verifying the proposed decision-making scheme on the three different fault cases, B1T2, B3T1x and B4T1x.

\subsection{Experimental Results and Discussion}

\begin{figure}[t]
\centering
\includegraphics[width=7cm]{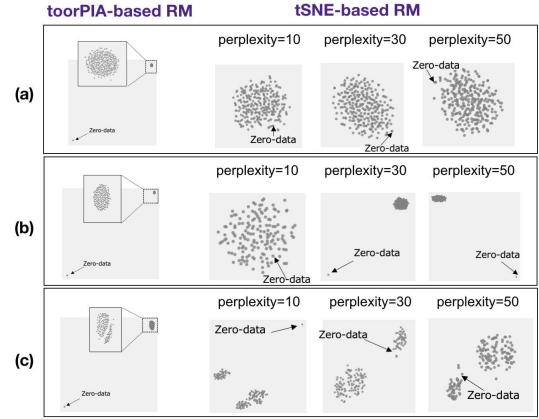}
\caption{The generated reference map from by toorPIA and t-SNE for (a) B1T2, (b) B3T1x and (c) B4T1x}
\label{fig14}
\end{figure}

\begin{figure}[t]
\centering
\includegraphics[width=7cm]{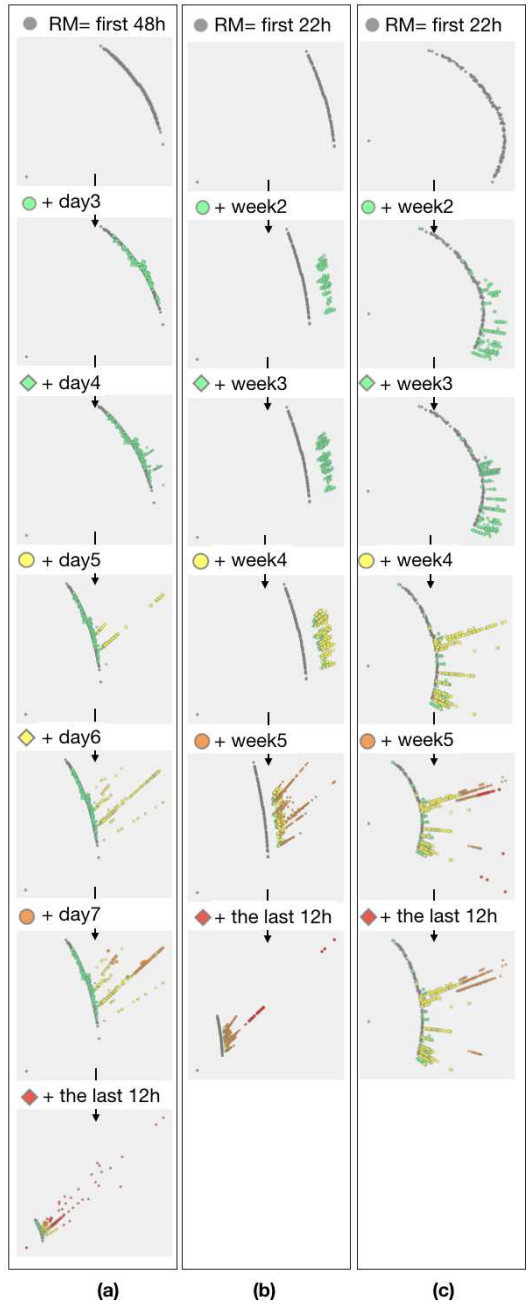}
\caption{The movement of the newly plotted data on the RM for (a) B1T2, (b) B3T1x and (c) B4T1x}
\label{fig15}
\end{figure}

\begin{figure}[t]
\centering
\includegraphics[width=7cm]{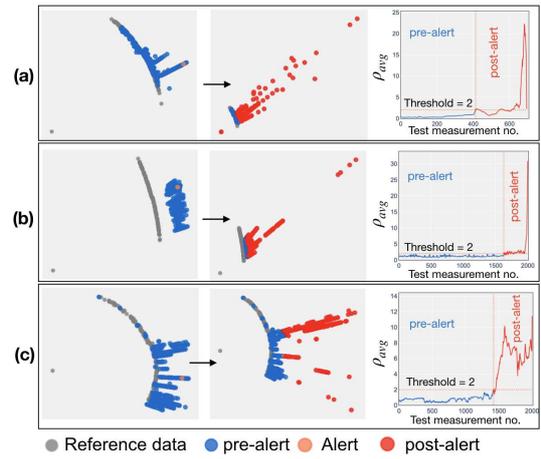}
\caption{Real-time tracking of B1T2 (a) before the alert, (b) after the alert}
\label{fig16}
\end{figure}

\begin{figure}[t]
\centering
\includegraphics[width=7cm]{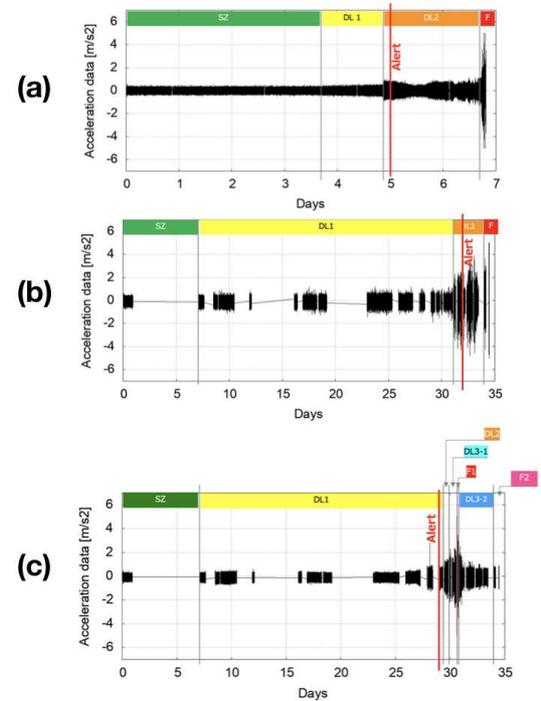}
\caption{Positioning of the RTDT alert in the fault time line given in section 5 (a) B1T2 case, (b) B3T1x and (c) B4T1x }
\label{fig19}
\end{figure}

The generated RM by toorPIA and t-SNE are shown by \textbf{Fig.\ref{fig14}.a} for B1T2, by \textbf{Fig.\ref{fig14}.b} for B3T1x and by \textbf{Fig.\ref{fig14}.c} for B4T1x. In the reference maps obtained by t-SNE, the \textit{zero-data} was sometimes positioned within the early data and sometimes far from the early data. This can be explained by the fact that t-SNE is not sensitive to far distance. On the other hand, toorPIA showed a stable differentiation between the \textit{zero-data} and the normal data among the three cases, B1T2, B3T1x and B4T1x. We are moving forward to present our results of plotting the test set using toorPIA-based RM. The use of t-SNE-based RM will be discussed in section 6.\par
\textbf{Fig.\ref{fig15}.a} illustrates how the test data was distributed daily for the B1T2 case. \textbf{Fig.\ref{fig15}.b} and \textbf{Fig.\ref{fig15}.c} present the weekly distribution of the test data for B3T1x and B4T1x respectively. The shape of the reference data as well as the position of the early added data in the two-dimensional space, when originated in the \textit{zero-data}, is reasonable because in the multidimensional space, normal data distribute in sphere around the \textit{zero-data}. Moreover, we can see, from \textbf{Fig.\ref{fig15}}, that as we progress in time, the test data gradually drift out of the reference data as well as from the normal data. Therefore, for the B1T2 case as well as for B3T1x and B4T1x cases, the RTDT successfully placed the test data on the RM in the right place, according the degradation level. However, if more data at the early stage was available, the test data will distribute in more natural way.\par
In \textbf{Fig.\ref{fig16}}, we are showing both the temporal movement of the newly plotted data in the two-dimensional plane with the graph of $\rho_{avg}$, for B1T2, B3T1x and B4T1x. In B1T2 (\textbf{Fig.\ref{fig16}.a}), an alert is to be issued on 02/17 at 11:12. In B3T1x (\textbf{Fig.\ref{fig16}.b}), an alert is to be issued on 11/23 at 04:26. As for B4T1x (\textbf{Fig.\ref{fig16}.c}), an alert is to be issued on 11/20 at 19:49. For all the three fault scenarios,  we can see that after the $\rho_{avg}$ reached $\rho_{avg_{th}}$, the data kept drifting far out of the safe zone.\par
In \textbf{Fig.\ref{fig19}} we are positioning the alert instance given by the RTDT in the timeline of the faults as we defined in the sub-section 4.3. We could verify the effectiveness of our proposed decision-making scheme. By using our proposal, we were able to issue an alert 2 days before the failure stage started in case B1T2 and B3T1x and 2 days before B4T1x exhibited the first type of failure. This proves that besides providing the operator with a 2D map to easily track the bearing condition from the data movement, we were able to guarantee enough time to the operator to conduct maintenance prior actual failure.

\section{Discussion}


\begin{table*}[t]
 \begin{center}
   \caption{Positioning of the proposed method among the related works}
   \resizebox{\textwidth}{!}{
   	\begin{tabular}[t]{|c|c|c|c|c|c|c|c|c|c|c} 
   	\hline
   	\multirow{2}{*}{Work} &Key  &Learning &Data  &Time & Frequency & Features &Diagnosis &Prognosis &Health \\
   	& technology\footnotemark & & visualization & domain signal\footnotemark[\value{footnote}] & domain signal\footnotemark[\value{footnote}] & & & &indicator\footnotemark[\value{footnote}]\\
   	\hline
   	Caesarendra et al. \cite{ref19}&LR+RVM&Yes&No&NU&EF&Yes&Yes&Limited&Failure Probability\\
   	Yu \cite{ref14}&LLP+NLLP&Yes&No&EF&EF&Yes&Yes&Limited&NLLPEWMA\\
   	Qiu et al. \cite{ref21}&WT+SOM&Yes&Yes&EF&De-noising&Yes&Yes&Limited&MQE\\
   	Mohamed et al. \cite{ref26}&Weibull hazard rates+ANN&Yes&No&EF&NU&Yes&Yes&Limited&Life percentage\\
   	Widodo and Yang \cite{ref29}&Survival analysis+RVM&Yes&No&EF&NU&Yes&Yes&Limited&Survival probability\\
   	Yu \cite{ref38}&GMM+k-means+SOM&Yes&No&EF&EF&Yes&Yes&Limited&NLLP and BIP\\
   	Yu \cite{ref36}&LNPP+KNN&Yes&No&EF&EF&Yes&Yes&Limited&H\\
   	Yu \cite{ref37}&DPCA+HMM&Yes&No&EF&EF&Yes&Yes&Limited&Mahalanobis distance\\
   	Dong and Luo \cite{ref22}&EMD+PCA+LSSVM&Yes&No&EF&EF&Yes&Yes&Limited&Extracted features values\\
   	Soualhi et al. \cite{ref24}&HMM+AAC&Yes&No&EF&NU&Yes&Yes&Limited&RTNDS\\
   	Qian et al. \cite{ref28}&PSW+Paris crack growth model&Yes&No&As vector&NU&No&Yes&Limited&RUL\\
   	Hasani et al. \cite{ref31}&AEC+MA&Yes&No&EF&NU&Yes&Yes&Limited&normalized AEC rate\\
   	Zhang et el. \cite{ref32}&WD+NB&Yes&No&EF&NU&Yes&Yes&Limited&Life percentage\\
   	Ben Ali et al. \cite{ref15}&EMD+ANN&Yes&No&EF&EF&Yes&Yes&Limited&HI\\
   	Our method&Full spectrum+data visualization&No&Yes&No&As vector&No&Yes&Full&WF($\rho$)\\
   	\hline
   	\end{tabular}
   	}
   \label{table6}
 \end{center}
\end{table*}
\footnotetext{LR: logistic Regression, RVM: Relevant Vector Machine, LLP: Local Preserved Projection, NLLP: Negative Log Likelihood Probability, WT: Wavelet Transform, SOM: Self Organized Map, ANN: Artificial Neural Network, GMM: Gaussian Mixture Models, LNPP: local and nonlocal preserving projection, KNN: K-nearest neighbor
DPCA: Dynamic Principal Components Analysis, HMM: Hidden Markov Model, EMD: Empirical Mode Decomposition, PCA: Principal Components Analysis, LS-SVM: Least-Squares Support-Vector Machine, AAC: Artificial ant clustering, PSW: Phase Space Wrapping, AEC: Auto Encoder Correlation, MA: Moving Average, WD: Weibull Distributions, NB: Naive Bayes, NLLPEWMA: Negative Log Likelihood Probability Exponentially Weighted Moving Average, MQE: minimum quantization error, BIP: Bayesian- inference, H: a Dynamic-Local-and-nonlocal-Preserving-Projection-based health index, RTNDS: Remaining time before the next degradation state, RUL: Remaining Useful Life, HI: Health Index, WF: Warning Factor, EF: Extracted Features, NU: Not Used}

\textbf{\textit{Comparison to existing methods:}}
\textbf{Table \ref{table6}} presents a comparison of the proposed solution to the works that have been published on the IMS dataset. If the key technology needs learning process and training data, the solution cannot be applied for a real-life failure prediction where the failures and faults have different patterns and unexpected timelines. Additionally, despite the effectiveness of the extracted features on the diagnosis of the bearing faults, a feature-based solution will limit the prognosis to a certain set of faults or degradation patterns that these features might express the best. Moreover, a solution based on data visualization is able to provide a more natural and human-friendly output. Therefore, compared to the cited works in \textbf{Table \ref{table6}}, we were able to fulfill the following: (1) No learning is required (2) No features extraction or selection (3) Use of data visualization and (4) Full prognosis \par

\textbf{\textit{Dynamic GUI as a real-time maintenance system:}}
In real life, our method is intended to be implemented and used in real-time as an RTDT with an appropriate Graphic User Interface (GUI). A reference map is made from the first one or two days of acquired data. Then, every new measurement can be plotted and added in the reference map using our proposed algorithm, in real-time. Additionally, the graph of $\rho_{avg}$ would be also updated in real-time with the acquired data. Therefore, the GUI would consist mainly of the 2D map as well as the $\rho_{avg}$ curve, as shown in \textbf{Fig.\ref{fig16}}. By visualizing on the GUI, dynamically, the movement of the newly plotted data points in the reference map and the evolution of the $\rho_{avg}$, the operator can monitor the condition of the machine and be ready to take the appropriate action in the appropriate time.\par
In this work, we used the data of the first 48h to generate a reference map then plotted the rest of the data into it as test data. If more data at the early stage is available, the test data will distribute in more natural way than the presented results in this paper. However, we are aware that the lifetime of the bearing might be short, and deterioration can happen at very early stage. Therefore, we propose the following two ideas: (i) Recording data more often in the first few hours of operation. Since the bearing are running continuously, we can collect data every 5 minutes for the first 24h. It is true that in the IMS dataset, in test 1, the first 43 measurements were taken every 5 minutes while the rest of the measurements were taken every 10 minutes. However, there were only extra measurement in the first 8h followed by a 6-days gap of measurements after the first 22h. (ii) If the first data that have been added to the reference map correspond to healthy data, we can update the reference map by including these data in the reference set.

\textbf{\textit{Possible t-SNE-based RTDT:}}
The current work used the toorPIA-based RM to be able to use the version of the plotting algorithm that was described in this work. However, as we stated in section 5.1, there are several approaches to plot the newly acquired data into the RM. Thus, t-SNE-based RM can be used with a modified version of the proposed plotting algorithm. t-SNE-based RM would be valid in the development of RTDT since based on the RTDT concept presented in the sub-section 5.1, while generating the RM, we do not require a high sensitivity to far distances.\par
In certain cases, such as chemical plants or heavy industry, machine failures can be fatal. The RTDT is effective in such situations where high precision in tracking machine condition and detecting early signs of failures is required to take the right action at the right time. The RTDT can be applied on other data types such as sound, chemical parameters and environmental parameters.

\section{Conclusion and Future Work}

This paper presented an innovative predictive maintenance scheme for machinery failure.	Our scheme combined the use of the full spectrum data as dimensions for the analysis instead of complicated time-based and/or frequency-based extracted features with the use of a visualization method. We were able to validate our proposed scheme on a real-world run-to-failure measurement provided by IMS bearing dataset. We started by verifying the ability of our proposed analysis scheme for the diagnosis of bearings. We then implemented our proposed scheme for a real-time data tracker as an online prognosis solution.\par
Our contributions are summarized as follows: (1) effectively classify and identify the different bearing conditions, (2) detect the early signs of degradation, (3) differentiate between types of failure, (4) RTDT as a real-time machine monitoring scheme for failure prediction and (5) no training data sets are required and the scheme can be used within a week after installation. \par
In the same research direction for proposing an RTDT as an on-line prognosis of machines/components, we are working on a second version of our plotting algorithm that can be used with t-SNE as visualization tool to generate the RM as discussed in section 6. Additionally, we are working on implementing a more accurate scheme for setting the alert based on the proposed warning factor $\rho$ as an index for degradation assessment.

\acknowledgements
The authors would like to thank Dr. Yoshio Takaeda the CTO of Toor Inc. for providing us with the license to use toorPIA as mapping engine.

\bibliographystyle{unsrt}
\bibliography{sanaTalmoudi_JRM.bib}

\begin{thebibliography}{10}

\bibitem{ref1}
Alberto Bellini, Fiorenzo Filippetti, Carla Tassoni, and G\'erard-Andr\'e
  Capolino.
\newblock Advances in diagnostic techniques for induction machines.
\newblock {\em IEEE Transactions on Industrial Electronics}, 55(12):4109--4126,
  2008.

\bibitem{ref2}
Austin~H. Bonnett and Chuck Yung.
\newblock Increased efficiency versus increased reliability.
\newblock {\em IEEE Industry Applications Magazine}, 14(1):29--36, 2008.

\bibitem{ref5}
Hai Qiu, Jay Lee, Jing Lin, and Gang Yu.
\newblock Wavelet filter-based weak signature detection method and its
  application on rolling element bearing prognostics.
\newblock {\em Journal of Sound and Vibration}, 289(4):1066 -- 1090, 2006.

\bibitem{ref6}
Jay Lee, Hai Qiu, Gang Yu, Jing Lin, and Rexnord Technical~Services (2007).
\newblock Bearing data set.
\newblock NASA Ames Prognostics Data Repository
  (http://ti.arc.nasa.gov/project/prognostic-data-repository).
\newblock IMS, University of Cincinnati, NASA Ames Research Center, Moffett
  Field, CA.

\bibitem{ref7}
Patrick Nectoux, Rafael Gouriveau, Kamal Medjaher, Emmanuel Ramasso, Brigitte
  Chebel-Morello, Noureddine Zerhouni, and Christophe Varnier.
\newblock Pronostia: An experimental platform for bearings accelerated
  degradation tests.
\newblock pages 1--8, 06 2012.

\bibitem{ref13}
Diego Fernandez-Francos, David Martinez, Oscar Fontenla-Romero, and Amparo
  Alonso-Betanzos.
\newblock Automatic bearing fault diagnosis based on one-class v-svm.
\newblock {\em Computers and Industrial Engineering}, 64:357–365, 01 2013.

\bibitem{ref23}
Haifeng Huang, Huajiang Ouyang, Hongli Gao, Liang Guo, Dan Li, and Juan Wen.
\newblock A feature extraction method for vibration signal of bearing incipient
  degradation.
\newblock {\em Measurement Science Review}, 16, 01 2016.

\bibitem{ref20}
Moises Diaz, Patricia Henríquez~Rodríguez, Miguel Ferrer, Giuseppe Pirlo,
  Jesús Alonso, Cristina Carmona-Duarte, and Donato Impedovo.
\newblock Stability-based system for bearing fault early detection.
\newblock {\em Expert Systems with Applications}, 79, 02 2017.

\bibitem{ref30}
Ran Zhang, Zhen Peng, Lifeng wu, Beibei Yao, and Yong Guan.
\newblock Fault diagnosis from raw sensor data using deep neural networks
  considering temporal coherence.
\newblock {\em Sensors}, 17:549, 03 2017.

\bibitem{ref33}
Rummaan Bin Amirand Sufi~Tabassum Gul and Abdul~Qayyum Khan.
\newblock A comparative analysis of classical and one class svm classifiers for
  machine fault detection using vibration signals.
\newblock In {\em 2016 International Conference on Emerging Technologies
  (ICET)}, pages 1--6, 2016.

\bibitem{ref17}
Yuyan Zhang, Xinyu Li, Liang Gao, and Peigen Li.
\newblock A new subset based deep feature learning method for intelligent fault
  diagnosis of bearing.
\newblock {\em Expert Systems with Applications}, 110, 06 2018.

\bibitem{ref18}
Toufik Berredjem and Mohamed Benidir.
\newblock Bearing faults diagnosis using fuzzy expert system relying on an
  improved range overlaps and similarity method.
\newblock {\em Expert Systems with Applications}, 108:134 -- 142, 2018.

\bibitem{ref34}
Sufi Gul, Munhal Imran, and Abdul Khan.
\newblock An online incremental support vector machine for fault diagnosis
  using vibration signature analysis.
\newblock pages 1467--1472, 02 2018.

\bibitem{ref16}
Ambika P~S, P.~Rajendrakumar, and R.~Ramchand.
\newblock Vibration signal based condition monitoring of mechanical equipment
  with scattering transform.
\newblock {\em Journal of Mechanical Science and Technology}, 33, 07 2019.

\bibitem{ref35}
Haisong Huang, Qingsong Fan, Jianan Wei, and Dong Huang.
\newblock An intelligent fault identification method of rolling bearings based
  on svm optimized by improved gwo.
\newblock {\em Systems Science and Control Engineering}, 7:289--303, 01 2019.

\bibitem{ref21}
Hai Qiu, Jay Lee, Jing Lin, and Gang Yu.
\newblock Robust performance degradation assessment methods for enhanced
  rolling element bearing prognostics.
\newblock {\em Advanced Engineering Informatics}, 17:127--140, 07 2003.

\bibitem{ref12}
Hai Qiu and Jay Lee.
\newblock Feature fusion and degradation using self-organizing map.
\newblock pages 107--114, 01 2004.

\bibitem{ref27}
Haitao Liao, Wenbiao Zhao, and Huairui Guo.
\newblock Predicting remaining useful life of an individual unit using
  proportional hazards model and logistic regression model.
\newblock pages 127 -- 132, 02 2006.

\bibitem{ref19}
Wahyu Caesarendra, Achmad Widodo, and Bo-Suk Yang.
\newblock Application of relevance vector machine and logistic regression for
  machine degradation assessment.
\newblock {\em Mechanical Systems and Signal Processing}, 24:1161--1171, 05
  2010.

\bibitem{ref26}
Abd Mahamad, Sharifah Saon, and Takashi Hiyama.
\newblock Predicting remaining useful life of rotating machinery based
  artificial neural network.
\newblock {\em Computers and Mathematics with Applications}, 60:1078--1087, 08
  2010.

\bibitem{ref14}
Jianbo Yu.
\newblock Bearing performance degradation assessment using locality preserving
  projections and gaussian mixture models.
\newblock {\em Mechanical Systems and Signal Processing}, 25(7):2573 -- 2588,
  2011.

\bibitem{ref29}
Achmad Widodo and Bo-Suk Yang.
\newblock Application of relevance vector machine and survival probability to
  machine degradation assessment.
\newblock {\em Expert Syst. Appl.}, 38(3):2592–2599, March 2011.

\bibitem{ref38}
Jianbo Yu.
\newblock A hybrid feature selection scheme and self-organizing map model for
  machine health assessment.
\newblock {\em Applied Soft Computing}, 11(5):4041 -- 4054, 2011.

\bibitem{ref36}
Jianbo Yu.
\newblock Local and nonlocal preserving projection for bearing defect
  classification and performance assessment.
\newblock {\em IEEE Transactions on Industrial Electronics}, 59(5):2363--2376,
  2012.

\bibitem{ref37}
Jianbo Yu.
\newblock Health condition monitoring of machines based on hidden markov model
  and contribution analysis.
\newblock {\em IEEE Transactions on Instrumentation and Measurement - IEEE
  TRANS INSTRUM MEAS}, 61:2200--2211, 08 2012.

\bibitem{ref22}
Shaojiang Dong and Tianhong Luo.
\newblock Bearing degradation process prediction based on the pca and optimized
  ls-svm model.
\newblock {\em Measurement}, 46:3143--3152, 11 2013.

\bibitem{ref15}
Jaouher Ben~Ali, N.~Fnaiech, Lotfi Saidi, Brigitte Chebel-Morello, and Farhat
  Fnaiech.
\newblock Application of empirical mode decomposition and artificial neural
  network for automatic bearing fault diagnosis based on vibration signals.
\newblock {\em Applied Acoustics}, 89:16–27, 03 2015.

\bibitem{ref24}
Abdenour Soualhi, Hubert Razik, Clerc Guy, and Dinh~Dong DOAN.
\newblock Prognosis of bearing failures using hidden markov models and the
  adaptive neuro-fuzzy inference system.
\newblock {\em IEEE Transactions on Industrial Electronics}, 61, 01 2013.

\bibitem{ref25}
Jaouher Ben~Ali, Brigitte Chebel-Morello, Lotfi Saidi, Simon Malinowski, and
  Farhat Fnaiech.
\newblock Accurate bearing remaining useful life prediction based on weibull
  distribution and artificial neural network.
\newblock {\em Mechanical Systems and Signal Processing}, 56, 05 2015.

\bibitem{ref28}
Yuning Qian, Ruqiang Yan, and Robert Gao.
\newblock A multi-time scale approach to remaining useful life prediction in
  rolling bearing.
\newblock {\em Mechanical Systems and Signal Processing}, 83, 07 2016.

\bibitem{ref31}
Ramin Hasani, Guodong Wang, and Radu Grosu.
\newblock An automated auto-encoder correlation-based health-monitoring and
  prognostic method for machine bearings.
\newblock {\em ArXiv}, 03 2017.

\bibitem{ref32}
Nannan Zhang, Lifeng wu, Zhonghua Wang, and Yong Guan.
\newblock Bearing remaining useful life prediction based on naive bayes and
  weibull distributions.
\newblock {\em Entropy}, 20:944, 12 2018.

\bibitem{ref10}
toor inc.
\newblock https://www.toor.jpn.com/.

\bibitem{ref11}
Laurens van~der Maaten and Geoffrey Hinton.
\newblock Visualizing data using t-sne.
\newblock {\em Journal of Machine Learning Research}, 9(86):2579--2605, 2008.

\bibitem{ref3}
Sana Talmoudi, Tetsuya Kanada, and Yasuhisa Hirata.
\newblock An iot-based failure prediction solution using machine sound data.
\newblock In {\em 2019 IEEE/SICE International Symposium on System Integration
  (SII)}, pages 227--232, 2019.

\bibitem{ref9}
Fabian Pedregosa, Ga{{\"e}}l Varoquaux, Alexandre Gramfort, Vincent Michel,
  Bertrand Thirion, Olivier Grisel, Mathieu Blondel, Peter Prettenhofer, Ron
  Weiss, Vincent Dubourg, Jake Vanderplas, Alexandre Passos, David Cournapeau,
  Matthieu Brucher, Matthieu Perrot, and {{\'E}}douard Duchesnay.
\newblock Scikit-learn: Machine learning in python.
\newblock {\em Journal of Machine Learning Research}, 12(85):2825--2830, 2011.

\end{thebibliography}


%

%
%
%

\appendix

\section{The RTDT proposed plotting algorithm}

For each test vector $\bm{X}$, we define $\bm{x}\in\Re^2$ where $\bm{x}=(x_x,x_y)$ as follow:\par
(1) In the multi-dimensional space, from the reference data, we find, $\bm{Z}$ as the closest data vector to $\bm{X}$ based on the multidimensional distance.\par
(2) In the multi-dimensional space, from the reference data, we find, $\bm{Y}$ forming the smallest angle with the data vector to $\bm{X}$.\par
(3) In both, the multi-dimensional space and the two-dimensional plane, we define $\bm{G}$, the center of gravity of the reference data.\par
(4) Using the geometrical relationship between $\bm{X}$, $\bm{Z}$, $\bm{Z}$, $\bm{O}$ and $\bm{G}$, we define the coordinates of $\bm{X}$ in the two-dimensional plane, in a coordinate system of the \textit{zero-data} as origin.\par
(5) All the data points of the reference map in the two-dimensional plane are then rotated as reference to $\bm{G}$.

\vspace{25pt}

%
%

\end{document}